\pdfoutput=1

\documentclass[11pt]{article}

\usepackage[]{EACL2023}

\usepackage{microtype}
\usepackage[utf8]{inputenc}
\usepackage[T1]{fontenc}
\usepackage{times}
\usepackage{latexsym}
\usepackage{amsfonts}
\usepackage{amsmath}
\usepackage{amssymb}
\usepackage{amsthm}
\usepackage{xspace}
\usepackage{float}

\usepackage{inconsolata} 
\usepackage{cleveref}
\usepackage[group-separator={,}]{siunitx}

\crefname{section}{\S}{\S\S}
\Crefname{section}{\S}{\S\S}
\crefname{table}{Tab.}{}
\crefname{figure}{Fig.}{}
\crefname{algorithm}{Algorithm}{}
\crefname{equation}{eq.}{}
\crefname{appendix}{App.}{}
\crefname{thm}{Theorem}{}
\crefname{prop}{Proposition}{}
\crefname{cor}{Corollary}{}
\crefname{observation}{Observation}{}
\crefname{assumption}{Assumption}{}
\crefformat{section}{\S#2#1#3}

\usepackage{tikz}
\usetikzlibrary{bayesnet}
\usetikzlibrary{arrows}
\usepackage{color}
\usepackage{graphicx}
\usepackage{caption}
\usepackage{subcaption}
\usetikzlibrary{backgrounds}

\usepackage{hhline}
\usepackage{xcolor}
\usepackage{colortbl}
\definecolor{Grey}{RGB}{210,210,210}

\usepackage{bbm}
\usepackage{todonotes}
\makeatletter
\newcommand*\iftodonotes{\if@todonotes@disabled\expandafter\@secondoftwo\else\expandafter\@firstoftwo\fi}  
\makeatother

\def\calE{{\mathcal{E}}}

\def\calT{{\mathcal{T}}}

\newcommand{\defn}[1]{\textit{#1}}

\newcommand{\OurApproach}{\text{PR-ENT}\xspace}
\newcommand{\PR}{\text{PR}\xspace}

\usepackage{tipa}

\newcommand{\linkprent}{https://huggingface.co/spaces/clef/PRENT-Demo}
\newcommand{\linkcodebook}{https://huggingface.co/spaces/clef/PRENT-Codebook}

\newcommand{\sF}{f}

\newcommand{\sY}{y}

\newcommand{\sE}{e}
\newcommand{\sT}{t}
\newcommand{\sTprime}{t'}

\newcommand{\sET}{\langle \sE, \sT \rangle}
\newcommand{\calET}{\mathcal{E} \times \mathcal{T}}

\newcommand{\sK}{k}
\newcommand{\bK}{K}

\newcommand{\sZ}{z_{\sE,\sT}}
\newcommand{\sZk}{z^{\sK}_{\sE,\sT}}
\newcommand{\sZstar}{z^{*}_{\sE,\sT}}

\newcommand{\bZ}{Z}

\newcommand{\bZstar}{Z^{*}}

\def\calY{{\mathcal{Y}}}

\def\calZ{{\mathcal{Z}}}
\def\calZt{{\mathcal{Z}_{t}}}
\def\calZk{{\mathcal{Z}^{\bK}_{e,t}}}
\def\calZstar{{\mathcal{Z}^{*}_{e,t}}}

\begin{document}

\title{Rethinking the Event Coding Pipeline with Prompt Entailment}

\author{Cl\'ement Lefebvre\Thanks{authors contributed equally}\\
  Swiss Data Science Center\\
  \normalsize\href{mailto:clement.lefebvre@datascience.ch}{\texttt{clement.lefebvre@datascience.ch}} \\\And
  Niklas Stoehr\footnotemark[1] \\
  ETH Z{\"u}rich\\
   \normalsize\href{mailto:niklas.stoehr@inf.ethz.ch}{\texttt{niklas.stoehr@inf.ethz.ch}}
}

\maketitle

\begin{abstract}
For monitoring crises, political events are extracted from the news. The large amount of unstructured full-text event descriptions makes a case-by-case analysis unmanageable, particularly for low-resource humanitarian aid organizations. This creates a demand to classify events into event types, a task referred to as event coding. Typically, domain experts craft an event type ontology, annotators label a large dataset and technical experts develop a supervised coding system. In this work, we propose \href{\linkprent}{\OurApproach}\footnote{\href{\linkprent}{\linkprent}}, a new event coding approach that is more flexible and resource-efficient, while maintaining competitive accuracy: first, we extend an event description such as ``Military injured two civilians'' by a template, e.g. ``People were [$\bZ$]'' and prompt a pre-trained (cloze) language model to fill the slot $\bZ$. Second, we select suitable answer candidates $\bZstar$ = \{``injured'', ``hurt''...\} by treating the event description as premise and the filled templates as hypothesis in a textual entailment task. In a final step, the selected answer candidate can be mapped to its corresponding event type. This allows domain experts to draft the codebook directly as labeled prompts and interpretable answer candidates. This human-in-the-loop process is guided by our \href{\linkcodebook}{codebook design tool}\footnote{\href{\linkcodebook}{\linkcodebook}}. We show that our approach is robust through several checks: perturbing the event description and prompt template, restricting the vocabulary and removing contextual information.
\end{abstract}


\section{Introduction}
\label{sec:intro}
Decision-makers in politics and humanitarian aid report a growing demand for comprehensive and structured overviews of socio-political events \cite{lepuschitz_seismographapi_2021}. For this purpose, news papers are automatically screened for event mentions, a task referred to as \defn{event detection} and \defn{extraction}. The sheer amount of extracted, full-text event descriptions day-to-day is impossible to be parsed by humans, especially when limited by scarce financial and computational resources.

\begin{figure}[t]
     \centering
     \includegraphics[width=1\linewidth]{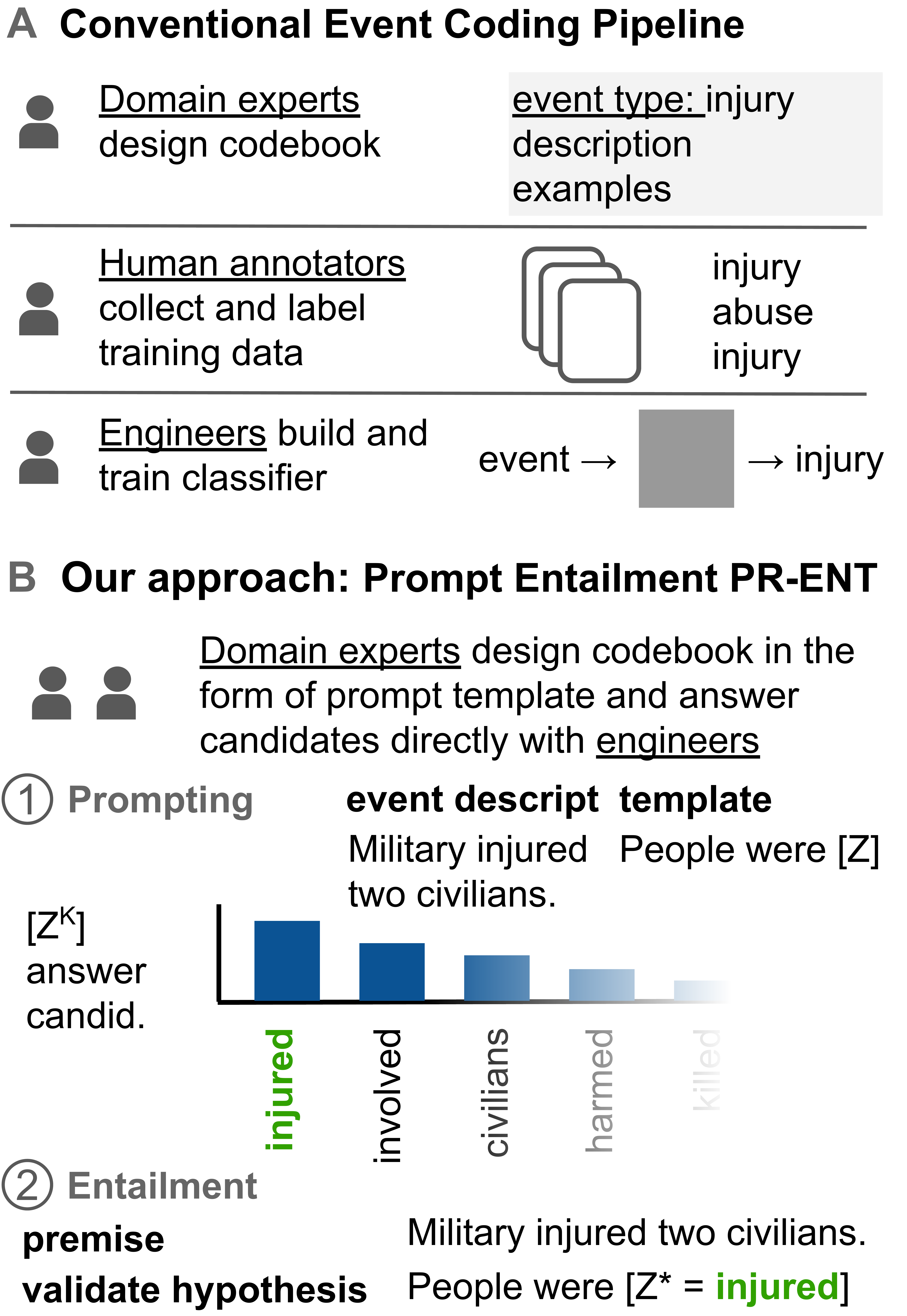} 
     \caption{\textbf{(A)} The conventional event coding pipeline involves many hand-overs between involved stakeholders and is strictly tailored to the event ontology. \textbf{(B)} Our approach combines prompting and textual entailment to perform flexible, unsupervised event coding.
     }
     \label{fig:overview}\label{fig:example}
 \end{figure}

\defn{Event coding} seeks to automatically classify event descriptions into pre-defined event types. Event coding is conventionally approached via a multi-step pipeline as shown in \cref{fig:overview}A. It incurs large costs in terms of human labor and time. We sketch out this pipeline expressed in \defn{human intelligence tasks (HITs)}\footnote{In our formulation, one HIT corresponds to roughly one hour of low-skill work by a single person such as reading and labeling single-sentence event descriptions. Our estimations are based on practical experience in working with domain experts and human annotators in the field of political event coding and serve the purpose of providing a very approximate quantification of required resources and labour.} \cite{ul_hassan_slua_2013}.

As a first step, an \defn{event ontology} is defined in terms of a codebook. Codebook development requires multiple domain experts \cite{goldstein_conflict-cooperation_1992} spending up to \num{200} HITs. The initial development phase of the widely-used \href{http://data.gdeltproject.org/documentation/CAMEO.Manual.1.1b3.pdf}{Conflict and Mediation Event Observations (CAMEO)} \cite{schrodt_cameo_2012} codebook reports a \num{3}-year initial development phase. Next, context-relevant event descriptions need to be collected to serve as training data. This often requires paid access to online newspaper distribution services and data collection infrastructure, estimated at \num{200} HITs. Next, human annotators need to be recruited and trained to annotate data according to the codebook accounting for another \num{200} HITs. Finally, a machine-based coding system needs to be developed, trained and validated, costing another \num{200} HITs. In earlier days, systems were dictionary- and pattern- based \cite{king_automated_2003, norris_petrarch2_2017}, while more recently machine learning-based approaches have gained momentum \cite{piskorski_tf-idf_2020, olsson_text_2020, hurriyetoglu_proceedings_2021}.

In total, the conventional event coding pipeline amounts to roughly \num{800} HITs. 
This development cost is often not bearable by non-profit / non-governmental organizations in the humanitarian aid sector. Moreover, the process requires multiple hand-overs between workers of different background which leads to errors, misunderstanding and delays. It is also important to highlight that the developed coding system is specifically tailored to a fixed event ontology. Any post-hoc changes of event types or even a different dataset incurs huge costs. In practice, event types frequently change and even vary widely between different divisions of the same organization.

To address these shortcomings, we present a new paradigm for highly adaptive event coding. Based on our method illustrated in \cref{fig:overview}B, domain experts are able to work directly with an interactive coding tool to design a codebook. They express event types by means of prompt templates and single-token answer candidates. For automated coding, a pre-trained language model is prompted to fill in those answer candidates taking a full-text event description as an input. Since prompting can be noisy \cite{gao_making_2021}, we propose filtering answer candidates based on textual entailment. Specifically, our contributions are as follows: (1) We propose a methodology combining prompting (\cref{sec:prompting}) and textual entailment (\cref{sec:entailment}) for event coding, termed \OurApproach. (2) We thoroughly evaluate this paradigm based on three aspects: accuracy (\cref{sec:accuracy}), flexibility (\cref{sec:flexibility}) and efficiency (\cref{sec:efficiency}). (3) We present two online dashboards: (a) A demo of the \href{\linkprent}{\OurApproach coding tool}. (b) An \href{\linkcodebook}{interactive codebook design tool} that guides the codebook design by presenting accuracy validation in a human-in-the-loop manner (\cref{sec:codebook_design}).

\section{Event Data and Types}
\label{sec:dataset_description}

We consider a subset of the \href{https://acleddata.com/#/dashboard}{Armed Conflict Location and Event Data (ACLED)} \cite{raleigh_introducing_2010} dataset. It is widely-used and has large coverage of political violence and protest events around the world. Each event is human annotated with a short description, its event type and additional details such as the number of fatalities and actor and targets. The event types are based on ACLED's own \href{https://acleddata.com/acleddatanew/wp-content/uploads/2021/11/ACLED_Codebook_v1_January-2021.pdf}{event ontology} which distinguishes \num{6} higher-level and \num{25} lower-level event types. Some event types are easily separable (e.g. \emph{protests} vs \emph{battles}), while others are harder to distinguish semantically (e.g. \emph{protests} vs \emph{riots}) (see \cref{fig:dataset_umap} in the appendix). 

We sample \num{4000} ACLED events (\num{3000} for training, \num{1000} for testing) in the African region while maintaining the event type distribution of the full dataset (see \cref{fig:dataset_umap}). We remove empty event descriptions and annotator notes (e.g. ``[size: no report]''). In \cref{fig:dataset_stats} in the appendix, we present statistics of the test set, showing different aspects of linguistic complexity. In \cref{sec:flexibility}, we consider the \href{https://www.start.umd.edu/gtd/}{Global Terrorism Dataset (GTD)} \cite{lafree_introducing_2007} to study the effect of domain shift. 

\section{Entailment-based Prompt Selection}
\label{sec:pipeline_description}
Our approach, \OurApproach, represents a real-world use case of prompting and textual entailment to code event descriptions $\sE \in \calE$ into event types $\sY \in \calY$ as shown in \cref{fig:overview}B.

\subsection{Prompting}
\label{sec:prompting}

\paragraph{Methodological Approach.}

In traditional supervised learning, a model is trained to learn a mapping between the input $\sE$ and the output class $\sY$. \defn{Prompting} \cite{liu_pre-train_2021} is a learning paradigm making use of (cloze) language models that have been trained to predict masked tokens within text.\footnote{``Cloze'' pertains to filling in missing tokens not necessarily uni-directional left-to-right, but anywhere in a string.} Prompt-based learning transfers this capability to perform classification in the following way:

We extend each \defn{event description} $\sE \in \calE$ by a \defn{template} $\sT \in \calT$ to form the input $\sET \in \calET$. Each template contains a \defn{masked slot} $\bZ$, e.g. ``This event involves [$\bZ$]'', ``People were [$\bZ$]''.\footnote{The first prompt template is intended to provide a one-word summary of the event. For the second template, we expect a verb describing the actions undertaken by the actor or a verb that describes what happened to the target.} The language model takes $\sET$ as input and returns an \defn{output distribution} of probabilities over the \defn{answer vocabulary} $\calZ$. Each token $\sZ \in \calZ$ can serve as a potential slot filler to $\bZ = \sZ$. However, we only consider the top $\sK$ most probable \defn{answer candidates} $\sZk \in \calZk$. $\calZ$ can be a constrained subset $\calZt$ that only features a template-related answer vocabulary to increase interpretability as pointed out in \cref{sec:ablation}. We discuss how to map answer candidates to event types in \cref{sec:accuracy}.

\paragraph{Implementation Details.}

We discuss the design of templates and constrained answer vocabularies resulting in a codebook (\cref{tab:ontology_from_prompt}) in \cref{sec:codebook_design}. In particular, we prompt \href{https://huggingface.co/distilbert-base-uncased}{DistilBERT-base-uncased} \cite{sanh_DistilBERT_2020}, a \defn{distilled} version of the BERT model which is more computationally efficient at the cost of a small performance decrease. For each prompt, we consider the $\bK = 30$ most probable tokens as the set of answer candidates $\calZk$. Ideally, we select a larger set, but performance gains are minimal while computational costs increase in subsequent steps. 

\begin{figure}[t]
     \centering
     \includegraphics[width=1\linewidth]{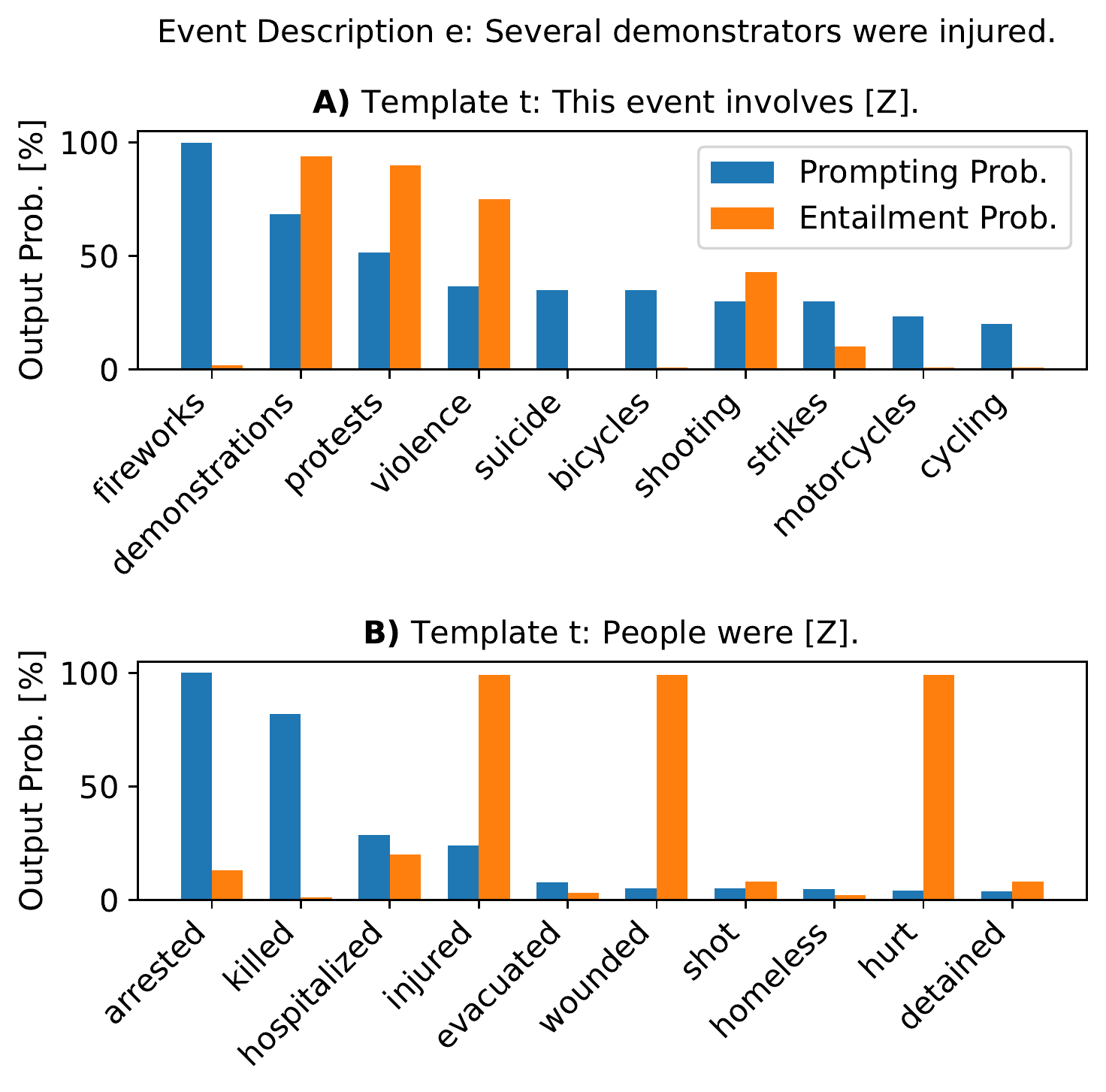} 
     \caption{Given the event description ``Several demonstrators were injured.'' and two templates \textbf{(A)} and \textbf{(B)}, prompting alone can yield tokens that fit syntactically but not semantically (blue bar). In contrast, filtering prompted answer candidates via textual entailment leaves us with tokens more closely related to the event (orange bar). To this end, we treat the event description as premise and the filled template as hypothesis. 
     }
     \label{fig:pnli_example}
 \end{figure}

\begin{table*}[t]
\fontsize{10}{10}\selectfont
\centering
\renewcommand{\arraystretch}{1.2} 
\setlength{\tabcolsep}{0.5em} 
\begin{tabular}{lll}
\textbf{Event Description + Template} $\sET$                                                                                                                                                   & \textbf{Answer Candidates}  $\sZk$                                                                                                                                          & \textbf{Entailed Answer Candidates} $\sZstar$                                                                       \\ \hline
\begin{tabular}[c]{@{}l@{}}Several demonstrators\\ were injured.\\ + People were {[}$\bZ${]}.\end{tabular}                                                         & \begin{tabular}[c]{@{}l@{}}arrested, killed, hospitalized,\\ injured, evacuated, \\ wounded, shot,\\ homeless, hurt, detained\end{tabular}                  & injured, wounded, hurt                                                                      \\ \hline
\begin{tabular}[c]{@{}l@{}}Several demonstrators\\ were injured.\\ + This event involves {[}$\bZ${]}.\end{tabular}                                                 & \begin{tabular}[c]{@{}l@{}}fireworks, demonstrations, \\ protests, violence, suicide, \\ bicycles, shooting, strikes,\\ motorcycles, cycling\end{tabular}   & \begin{tabular}[c]{@{}l@{}}demonstrations, protests, \\ violence\end{tabular}               \\ \hline
\begin{tabular}[c]{@{}l@{}}The sponsorship deal\\ between the shoes brand\\ and the soccer team\\ was confirmed.\\ + This event involves {[}$\bZ${]}.\end{tabular} & \begin{tabular}[c]{@{}l@{}}sponsorship, nike, sponsors, \\ fundraising, cycling, \\ advertising, charity, donations, \\ concerts, competitions\end{tabular} & \begin{tabular}[c]{@{}l@{}}sponsorship, sponsors, \\ advertising, competitions\end{tabular} \\ \hline
\end{tabular}
\caption{We prompt a language model based on an event description $\sE$ and template $\sT$ with slot $\bZ$. We keep only those prompted answer candidates $\sZk \in \calZk$ entailed in a subsequent textual entailment task $\sZstar \in \calZstar$.}
\label{tab:prompt_nli_example}
\end{table*}

\subsection{Textual Entailment}
\label{sec:entailment}

\paragraph{Limitations of Prompting.}

Prompting yields event-related tokens for event coding, but comes with challenges. There is no guarantee that a prompted answer candidate $\sZk \in \calZk$ is suited to represent an event. Answer candidates may be semantically unrelated as shown in \cref{fig:pnli_example}. To address this shortcoming, we propose filtering $\calZk$ via textual entailment. Textual entailment, or natural language inference (NLI) \cite{fyodorov_natural_2000, bowman_large_2015} can be framed as the following task: Given a ``premise'', verify whether a ``hypothesis'' is true (entailment), false (contradiction), or undetermined (neutral). It has been evaluated as a popular method for performing text classification \cite{wang_entailment_2021}.

\paragraph{Selecting Entailed Answer Candidates.}

We consider the event description $\sE$ as premise and the template $\sTprime$ filled with a prompted answer candidate as hypothesis. For example, given the premise ``Two bombs detonated...'', we automatically construct hypotheses ``This event involves $[\sZk] \in \calZk = \{ \text{explosives, civilians...}\}$'', see \cref{tab:prompt_nli_example}. We pass the concatenation of the premise and hypothesis to \href{https://huggingface.co/roberta-large-mnli}{RoBERTa-large-mnli} \cite{liu_RoBERTa_2019}. If the model finds premise and hypothesis to be entailed, then the prompted answer candidate $\sZk$ is considered an \defn{entailed answer candidate} $\sZstar$ (e.g. $\sZstar$ = explosives). We combine the categories ``neutral'' and ``contradiction'' into one since we are interested in a hypothesis being entailed or not.

\begin{figure}[h]
    \centering
    \includegraphics[width=0.95\linewidth]{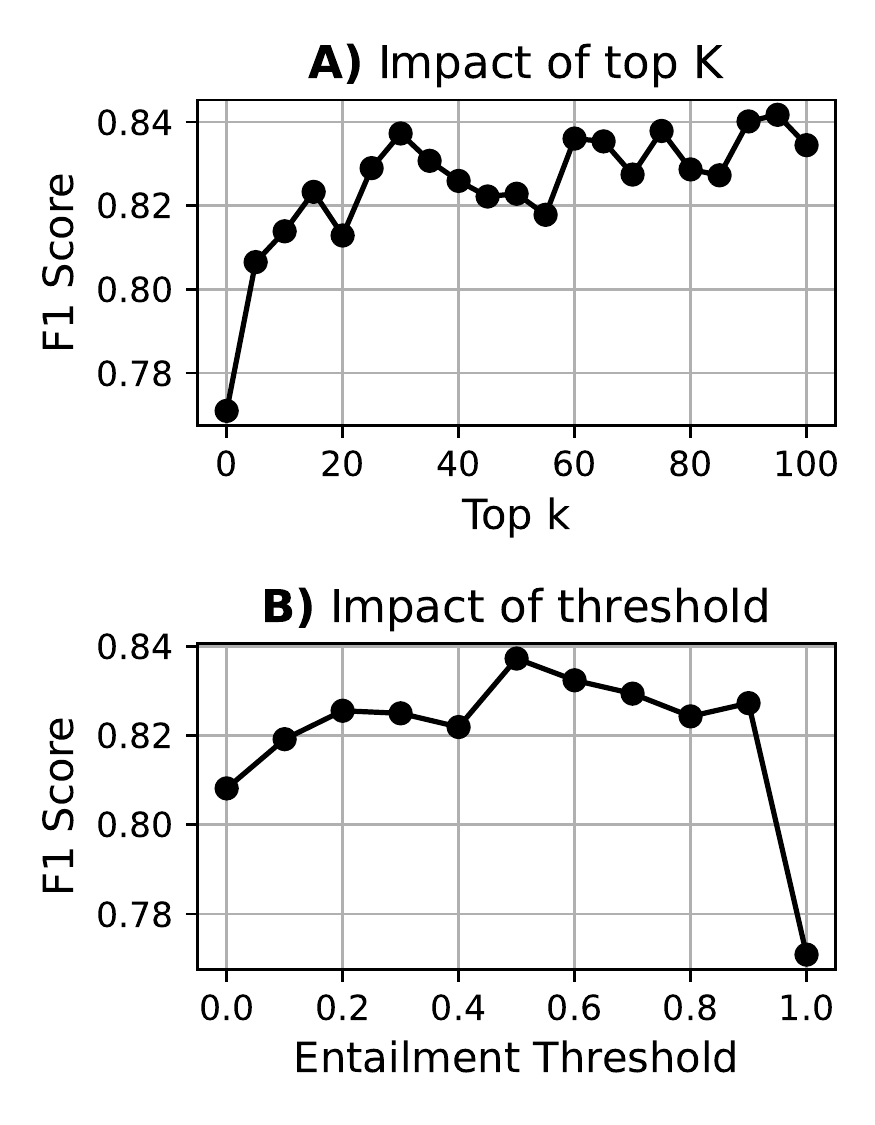} 
    \caption{Impact of different parameters of our pipeline on ACLED classification. \textbf{(A)} F1 score versus the maximum number $\bK$ of allowed answer candidates; $\bK = 0$ means that only the event description is used in the classification. \textbf{(B)} F1 score versus entailment threshold; the threshold governs if a hypothesis is entailed with the premise or not, a threshold of $0$ means that all prompted answer candidates are considered. A threshold of $1$ means only the event description is considered.
    }
    \label{fig:top_k_nli_threshold}
\end{figure}

This means, \OurApproach has two hyperparameters: the top $K$ answer candidate tokens yielded by the prompting step and the acceptance threshold in the entailment step that governs whether an answer candidate is kept. We empirically analyse the effect of both hyperparameters on the final F1 classification score in \cref{fig:top_k_nli_threshold}. In \cref{fig:top_k_nli_threshold}A, we verify that considering the top \num{30} answer candidate tokens leads to good performance on average. Further, we find a suitable threshold of \num{0.5} on the entailment model's output probability in \cref{fig:top_k_nli_threshold}B.


\section{Evaluation: Event Classification}
\label{sec:evaluation}

We compare \OurApproach against the conventional event coding pipeline in an evaluation along three dimensions: accuracy, flexibility and efficiency. 

\subsection{Accuracy}
\label{sec:accuracy}

So far we have not discussed how to map entailed answer candidates $\sZstar \in \calZstar$ onto event types $\sY \in \calY$. We choose to do \defn{hard} prompting, as opposed to \defn{soft} prompting. This means, tokens in $\calZstar$ are mapped onto event types $\sY$ via a simple linear transform $\sY = \sF(\sZstar)$. When $\sF$ is the identity function, no additional mapping is needed (\cref{sec:flexibility}). Hard prompting allows defining event types, i.e. an event ontology, in terms of interpretable answer candidates. As an example, we present an interpretable event ontology in \cref{tab:ontology_from_prompt} in the appendix. We use it to classify ``lethal'' and ``non-lethal'' event as explained in \cref{sec:flexibility}. Generally, we observe a trade-off between accuracy and interpretability. We want different sets of entailed answer candidates to uniquely define different event types at a high accuracy. At the same time, we require the set to be limited to a few, interpretable tokens only, that are highly representative for the event type. In the following, we learn a shallow mapping between $\calZstar$ and the \num{6} high-level event types $\calY$ provided by the \href{https://acleddata.com/acleddatanew/wp-content/uploads/2021/11/ACLED_Codebook_v1_January-2021.pdf}{ACLED event ontology} as ground truth.

\paragraph{Baselines and Ceilings.}
As baselines, we consider \defn{bag-of-words} (BoW) and GloVe \cite{pennington_glove_2014} embeddings of event descriptions. Embeddings are mapped onto event types via logistic regression (LR). Further, we contrast our \OurApproach with a prompting-only (\PR) approach also using logistic regression as a classification layer. As a ceiling model, we consider DistilBERT fine-tuned in a sequence classification task.

\paragraph{Our Approach \OurApproach.}

To evaluate our approach, we only consider the template ``This event involves [$\bZ$]'' and construct a BoW feature matrix by extending the event descriptions $\sE$ with the entailed answer candidates $\sZstar$. The resulting feature matrix serves as input to logistic regression. We report classification results in \cref{tab:acc_f1_results} and find that \OurApproach is only outperformed by the supervised, fine-tuned DistilBERT ceiling, but performs better than all baselines.

\begin{table}
\fontsize{10}{10}\selectfont
\centering
\renewcommand{\arraystretch}{1.45} 
\setlength{\tabcolsep}{0.3em} 
\begin{tabular}{lll}
\textbf{Model}                                  & \textbf{Accuracy}       & \textbf{F1 Score}      \\ \hline
\multicolumn{1}{l|}{BoW + LR}          & 80.5          & 77.1         \\
\multicolumn{1}{l|}{GloVe + LR}        & 78.5          & 74.6         \\
\multicolumn{1}{l|}{Random Tokens + BoW + LR} & 77.1        & 72.2         \\
\multicolumn{1}{l|}{PR + BoW + LR} & 82.9        & 80.8         \\
\multicolumn{1}{l|}{\OurApproach + BoW + LR} & 85.1          & 83.7         \\
\multicolumn{1}{l|}{DistilBERT}        & \textbf{87.1} & \textbf{86.0}
\end{tabular}
\caption{Classification of \num{6} event types in the ACLED dataset. As expected, DistilBERT performs best as it is fine-tuned specifically on this classification task. Our approach \OurApproach is more ad-hoc and does not fall far behind. The additional entailment step reduces noise compared to the prompting-only approach \PR. On top of the two standard baselines using BoW and GloVe, we introduce an additional baseline where we select \num{10} random tokens from $\calZk$ for each $\sET$. Compared to all baselines, \OurApproach performs better.}
\label{tab:acc_f1_results}
\end{table}

 \begin{figure*}[h]
     \centering
     \includegraphics[width=0.95\linewidth]{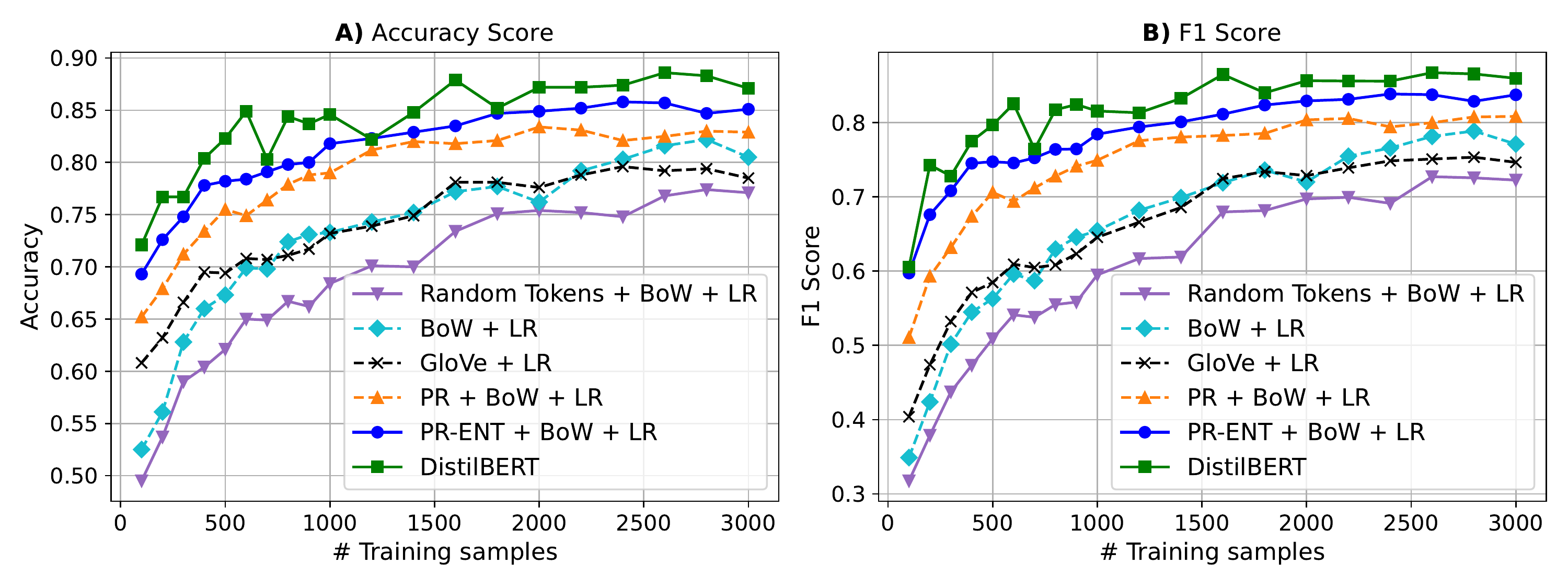} 
     \caption{Comparison of the different classification approaches on a varying number of training instances. Our approach \OurApproach shows better performance in terms of accuracy and F1 Score than the baseline models at all points. At the same time, it does not lack far behind the fine-tuned DistilBERT ceiling model, which is however less flexible and resource-intensive. \PR refers to prompting-only, BoW to bag-of-words and LR to logistic regression. The baseline ``random'' consists of sampling \num{10} random tokens from $\calZk$ for each $\sET$.}
     \label{fig:training_instances_evo}
 \end{figure*}

\subsection{Flexibility}
\label{sec:flexibility}

We explore the flexibility of \OurApproach along \num{3} dimensions: changing the number of training instances, omitting the shallow mapping for classification and switching to another dataset.

\paragraph{Number of Training Instances.}

As can be seen in \cref{fig:training_instances_evo}, our approach shines at classifying event types if only few training instances are given. \OurApproach shows better performance than all baseline approaches introduced in \cref{sec:accuracy}. At the same time, it is not far behind the fine-tuned DistilBERT ceiling model.

\paragraph{Removing the Shallow Mapping.}

We may remove the requirement of adding a shallow mapping $\sY = f(\sZstar)$. Therefore, we predict if an event is ``lethal'' ($\sY = 1$) or not ($\sY = 0$) based on its description. We use \OurApproach to generate entailed answer candidates $\calZstar$ based on the template ``People were [$\bZ$].''. If $\bZ = \text{``killed''} \in \calZstar$ then $\sY = 1$. We compare \OurApproach against fine-tuned DistilBERT trained on \num{100} samples and present results in \cref{tab:lethal_event_results}. \OurApproach is competitive against DistilBERT, even outperforming it in this setting. Moreover, while the prompting-only approach (\PR) has very high recall, it lacks precision. The additional entailment step in \OurApproach balanced this out, yielding a high F1 score.

\begin{table}
\fontsize{10}{10}\selectfont
\centering
\renewcommand{\arraystretch}{1.4} 
\setlength{\tabcolsep}{0.5em} 
\begin{tabular}{llll}
\textbf{Model}                                  & \textbf{F1 Score} & \textbf{Precision} & \textbf{Recall} \\ \hline
\multicolumn{1}{l|}{\OurApproach}             & \textbf{91.6}    & \textbf{85.3}     & 98.8           \\
\multicolumn{1}{l|}{Prompting Only}        & 50.6               & 33.9              & \textbf{100}  \\
\multicolumn{1}{l|}{DistilBERT} & 84.1             & 76.5              & 93.4           \\
\end{tabular}
\caption{Binary classification of ``non-lethal versus lethal'' events based on ACLED's fatality counts. In \OurApproach and prompting-only \PR, we code ``lethal'' if ``killed'' is present in the answer candidates of ``People were [$\bZ$].''. We observe the added value of the entailment step in the increase in precision. \OurApproach outperforms DistilBERT trained on \num{100} data instances and tested on \num{1000} event descriptions.}
\label{tab:lethal_event_results}
\end{table}

\paragraph{Domain Shift.} We scrutinize the robustness of \OurApproach by switching to another dataset. We repeat the binary ``lethal versus non-lethal'' classification task on the \emph{Global Terrorism Database} (GTD) \cite{lafree_introducing_2007}. The results in \cref{tab:gtd_lethal_event_results}, again suggest high performance of \OurApproach.

\begin{table}[h]
\fontsize{10}{10}\selectfont
\centering
\renewcommand{\arraystretch}{1.4} 
\setlength{\tabcolsep}{0.5em} 
\begin{tabular}{llll}
\textbf{Model}                            & \textbf{F1 Score} & \textbf{Precision} & \textbf{Recall} \\ \hline
\multicolumn{1}{l|}{\OurApproach}     & \textbf{96.3}    & \textbf{94.0}      & 98.8           \\
\multicolumn{1}{l|}{Prompting Only}  & 67.3             & 50.7              & \textbf{100}      \\
\multicolumn{1}{l|}{DistilBERT}  & 93.4               & 89.9                & 97.2             \\
\end{tabular}
\caption{Binary classification of ``non-lethal versus lethal'' based on the Global Terrorism Database (GTD). \OurApproach and prompting-only \PR predict ``lethal'' if ``killed'' is prompted from ``People were [$\bZ$].''. \OurApproach outperforms DistilBERT trained on 100 data instances and tested on \num{1000} event descriptions.}
\label{tab:gtd_lethal_event_results}
\end{table}

\begin{table*}[t]
\fontsize{10}{10}\selectfont
\centering
\renewcommand{\arraystretch}{1.57} 
\setlength{\tabcolsep}{0.5em} 
\begin{tabular}{l|llllllll}
\textbf{Perturbation Type} & \multicolumn{2}{l}{\textbf{Paraphrase}} & \multicolumn{2}{l}{\textbf{Remove Stop Words}} & \multicolumn{2}{l}{\textbf{Remove Entities}} & \multicolumn{2}{l}{\textbf{Duplication}} \\ \hline
Model Type        & PR        & \OurApproach              & PR        & \OurApproach              & PR        & \OurApproach             & PR         & \OurApproach              \\
1 Perturbation    & 0.33      & \textbf{0.14}      & 0.22      & \textbf{0.15}      & 0.15      & \textbf{0.08}     & 0.18       & \textbf{0.09}      \\
2 Perturbations   & 0.34      & \textbf{0.18}      & -         & -                  & -         & -                 & 0.28       & \textbf{0.16}     
\end{tabular}
\caption{Average Jensen-Shannon distance across \num{1000} event descriptions. We conduct \num{4} perturbation tests: paraphrasing the template, removing stop words from the event description, replacing named entities by a placeholder, and duplicating words in the template. \OurApproach is more robust than \PR: in all cases, the distance between the output distributions based on the non-perturbed and perturbed input is smaller.}
\label{tab:perturbations}
\end{table*}

 \begin{figure*}[t]
     \centering
     \includegraphics[width=0.95\linewidth]{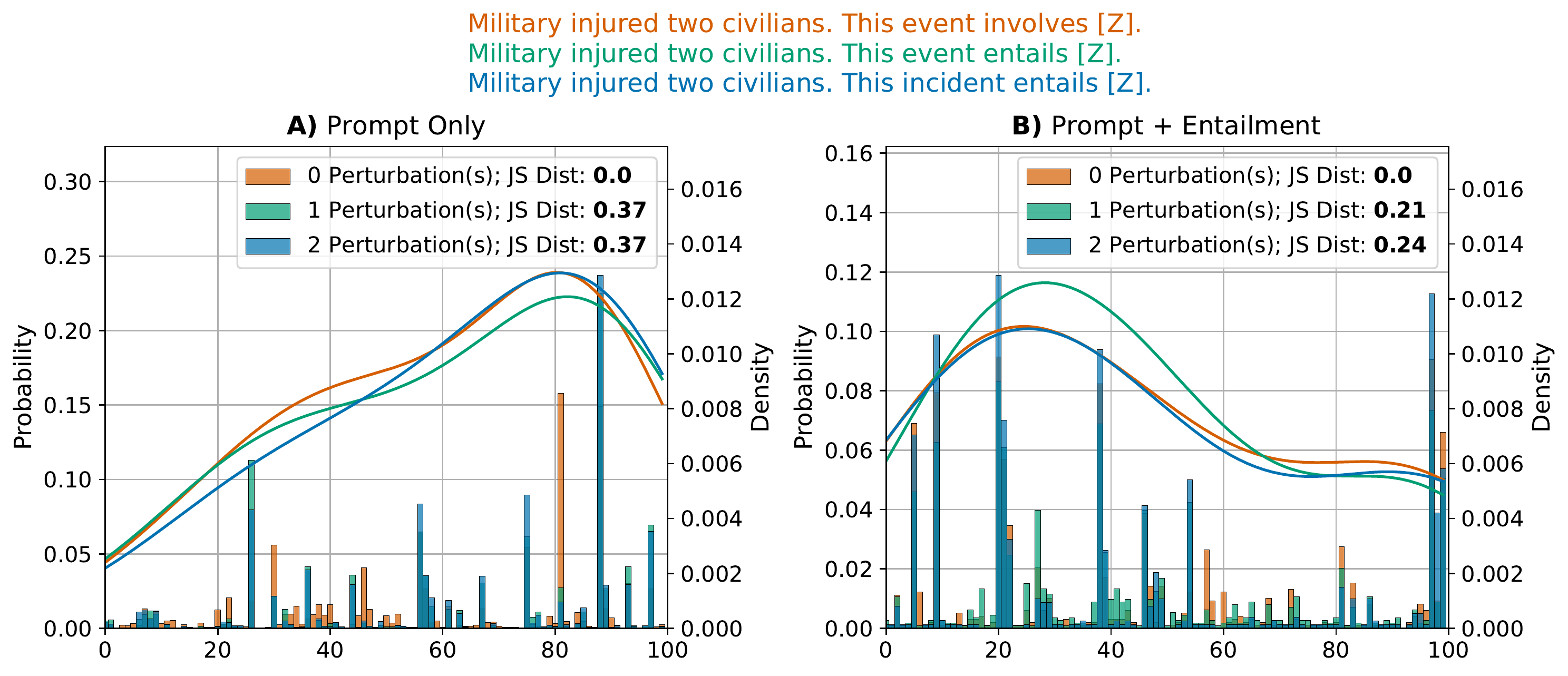} 
     \caption{We compare prompting-only \PR and our approach \OurApproach when perturbing the input $\sET$. \OurApproach is more robust to perturbations as indicated by a lower Jensen-Shannon distance between the output distributions over answer candidates based on non-perturbed and perturbed input. \PR is highly sensitive to template phrasing. X-label represents the top 100 most frequent tokens from 1000 prompts.}
     \label{fig:perturbation_prompt_example}
 \end{figure*}

\subsection{Efficiency}
\label{sec:efficiency}

In \cref{sec:intro}, we estimated the cost of \num{800} human intelligence tasks (HIT) for the conventional event coding pipeline. We perform the same estimation exercise for our approach: domain experts design suitable templates and answer candidate sets in a trial and error fashion as elaborated in \cref{sec:codebook_design}. We estimate total development costs at about \num{300} HITs, which makes it particularly feasible for small teams with few resources such as non-governmental organizations in humanitarian aid. Overall, our approach requires fewer people and consequently fewer hand-overs. Moreover, it is not tied to a specific event ontology and more flexible for changing event types.

\section{Ablation Study}
\label{sec:ablation}

\subsection{Perturbation Tests}

Our approach is not tailored to a specific event ontology, but to a language model. Any performance gains on these models, such as the recently published ConfliBERT \cite{hu_conflibert_2022}, will impact our pipeline. A crucial consideration is the presence of biases within language models. In some settings, biases may even be desirable inductive priors, but should at least be known. 

We measure the sensitivity of the prompted model's output distribution to changes in the input. To this end: we select a fixed answer vocabulary $\calZt$ of \num{100} tokens by taking the most frequent tokens yielded by the prompted model across \num{1000} event descriptions. We observe the output distribution over tokens in $\calZt$ before and after perturbing the input $\sET$. Finally, we measure the difference between the two output distributions in terms of \href{https://docs.scipy.org/doc/scipy/reference/generated/scipy.spatial.distance.jensenshannon.html}{Jensen-Shannon (JS) distance}. We show the results of the following four perturbation settings in \cref{tab:perturbations}:

\vspace{0.2cm} 
\noindent\textbf{(1) Paraphrasing} Two prompt designers could come up with paraphrased templates. In \cref{fig:perturbation_prompt_example}, we show that the additional entailment step makes \OurApproach more robust to perturbations in the template as opposed to prompting only.

\noindent\textbf{(2) Stop Word Removal} We remove stop words from the event description to test \OurApproach on non-grammatical text.

\noindent\textbf{(3) Context Removal} We remove all named entities in event descriptions and replace them with placeholder tokens such as ``organizations'' and ``locations''. This verifies that \OurApproach is less prone to latching onto context instead of content.

\noindent\textbf{(4) Duplication} We duplicate some words in the template. Specifically, we test the \num{3} prompts: ``This event involves [$\bZ$]'', ``This event event involves [$\bZ$]'', ``This event event event involves [$\bZ$]''.

\subsection{Comparing Coded Event Time Series}

\begin{figure*}[t]
 \centering
 \includegraphics[width=1\linewidth]{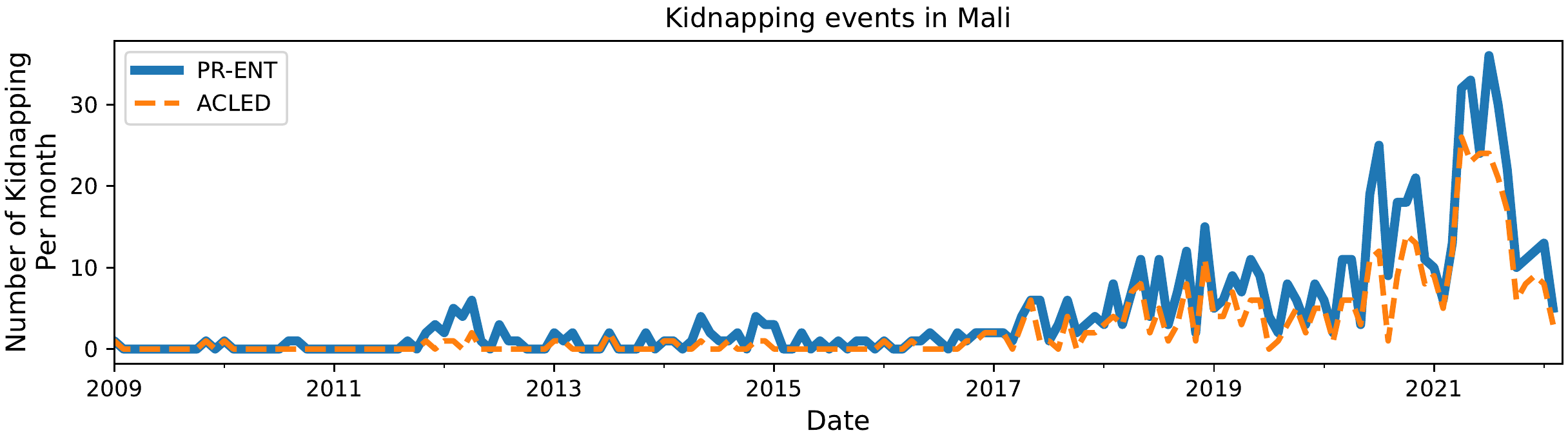} 
 \caption{Time series of the number of kidnapping events per month in Mali between 2009-2021. The dashed line corresponds to all kidnapping events coded by ACLED annotators. The blue line corresponds to all kidnapping events coded by \OurApproach. We find that the positions of the time series spikes between \OurApproach and ACLED's coding align well. However, the spikes in the \OurApproach time series are higher indicating that \OurApproach detects more events. This may be due to more granular event coding or the advantage of not being limited to only one event type per event description.}
 \label{fig:mali_kidnappings}
\end{figure*}

Using \OurApproach, we construct a codebook (\cref{tab:ontology_from_prompt}) to code ACLED event descriptions without the need of a shallow mapping. We use this codebook to code events that took place in Mali (\cref{fig:mali_kidnappings}) and Ethiopia (\cref{fig:mali_kidnappings}) between 2009 and 2021. This allows comparing time series of event types between our approach and ACLED's coding. We find that both codings yield very similar time series in which the positioning of spikes align. Yet, the spikes in the \OurApproach time series are higher / steeper indicating that more events are detected. This may be attributed to two reasons: firstly, \OurApproach is potentially more granular and has higher recall. Secondly, \OurApproach is not limited to coding only one event type per event description as ACLED is. For example, the following event description in ACLED (anonymized) is coded as Armed Clash but contains several possible event types (Armed Clash, Killing, Kidnapping, Property Destruction, Looting): ``[...] The militants clashed with [ORG], and killed one [ORG] and a civilian driver, abducted one person, burned a vehicle and seized livestock.''

\subsection{Qualitative Error Analysis}

We perform a qualitative error analysis of our proposed method. Within the ACLED data, there are many event descriptions containing mentions of past events (e.g. ``Protests over the killing of the journalist [NAME] shot dead on Monday at his home by armed bandits.''). Our method, and in fact, any supervised classifier, may have difficulties recognizing event co-references. Another frequent error is due to ACLED event type definitions. For instance, ACLED features the event type ``Violence Against Civilians''. However, to classify most of the concerned events, the annotator needs to know if the target is a civilian or not. Unfortunately, the dataset does not always contain this information, except if explicitly written in the event description. Another frequently observed error is caused by blurry definition of event types. ACLED, differentiates between ``Riots'' and ``Protests'' which often have nearly identical event descriptions.

\section{Human-Computer Codebook Design}
\label{sec:codebook_design}

To make use of \OurApproach, domain experts need to design a codebook (i.e. a mapping), between event types and entailed answer candidates. Creating this mapping is non-trivial as there exists a trade-off between interpretability and accuracy. In essence, a codebook is interpretable when the answer candidates are representative of the corresponding event type. A bad codebook contains a large number of non-readable entailed answer candidates. A codebook is accurate when a few answer candidates are sufficient to allow for a clear differentiation of the event types. To that end, we propose an \href{\linkcodebook}{interactive codebook design tool}\footnote{\href{\linkcodebook}{\linkcodebook}} that helps designing templates and answer candidates by presenting accuracy metrics. The assessment of interpretability is left to the human domain experts. 

\paragraph{Codebook Design.} Our codebook is a mapping between event types and entailed answer candidates. For example, an event can be classified as ``kidnapping'' if any of the following templates is entailed: ``This event involves [kidnapping].'' \texttt{OR} ``This event involves [abduction].''. A codebook example is shown in \cref{tab:ontology_from_prompt} in the appendix. 

We assume two things: first, the availability of a dataset which contains event descriptions that need to be labeled. Second, the domain experts should have decided upon event types of their liking (e.g. kidnapping, killings,...). Now, the first step is to come up with an initial set of templates and entailed answer candidates. For each event type, the domain expert is asked to draft a canonical event description. For example: the event type ``kidnapping'' could be exemplified by ``Two men were kidnapped by rebels.''. Then using \OurApproach, the domain expert is presented a list of answer candidates (e.g. ``This event involves [kidnapping].'', ``This event involves [rebels].''...). 

As a second step, domain experts select some of the entailed answer candidates provided by the model. If no entailed answer candidate is informative to classify the event, it is possible to group multiple entailed answer candidates with an \texttt{AND} condition. For example, ``Riot'' event types can be coded with the two following templates: ``This event involves [protest].'' \texttt{AND} ``This event involves [violence].''. The tool also offers the possibility of excluding certain answer candidates.

\paragraph{On-the-Go Validation.}
Validating the interpretability of the codebook and the answer candidates is a subjective task that we leave to the domain experts. The coding tools offers however guidance for the validation of accuracy, despite not having access to ground truth event type labels. Using the current state of the codebook and \OurApproach, randomly selected events are automatically coded into event types. Domain experts can then accept or reject the event type suggestions provided by the model. This creates a labeled dataset ``on the go'', which allows computing a per-class accuracy score. Repeated rounds of validation allow for a human-in-the-loop fine-tuning of the codebook by adding or removing more entailed answer candidates.

\paragraph{Codebook Use.}
The tool offers interoperability by enabling the download of the codebook and the labeled dataset in standard JSON format. The former can then be used to code a full dataset of event descriptions into event types. The codebook can still be modified if more event types are required.

\section{Discussion}
\label{sec:limitations}

\paragraph{Is this few-shot, unsupervised tagging?}

While we have evaluated accuracy, efficiency and flexibility, it is up for discussion and definition whether our approach should be considered few-shot, unsupervised or tagging-based. In some cases, the language model copies tokens verbatim from the input, which could be seen as a form of ``event tagging''. In other cases, the answer candidates are abstract tokens outperforming purely tagging-based approaches. In cases where the answer candidates map directly to an event type without an additional shallow classifier \cref{sec:flexibility}, our approach may be considered unsupervised and zero-shot. On the contrary, the template is designed in an iterative trial and error fashion. Thus, it is tuned to observed data instances which arguably violates the zero-shot setting and should be framed few-shot instead.

\paragraph{Entailment-Only Approach.}

The presented approach \OurApproach relies on textual entailment to select entailed answer candidates from prompts as motivated in \cref{sec:entailment}. However, textual entailment could have been considered for classification by itself \cite{wang_entailment_2021, barker_ibm_2021}. In this setting: a predefined set of hypotheses is created for each event type and is tested against each event description. However, this reduces flexibility as we need to define a broad set of hypotheses in advance. Our prompting-based approach relies on large language models which do not require labeled training data for training. As a consequence, they are more frequently updated and trained on larger amounts of data.  

\paragraph{Extensions and Applications.}

Our approach can be used to filter and search events in a dataset of full-text event descriptions. An example of this use case is described in \cref{sec:flexibility} where we classify lethal and non-lethal events in an unsupervised way via the ``killed'' token. Promising extension are the coding of source and target actors in addition to event types as presented in \cref{sec:actor_target_coding} as well as the extraction of victim counts \cite{zhong_extracting_2023}.

\section{Related Work}

Similar to our prompting-based approach, existing work evaluates off-the-shelf QA \citep{halterman_corpus-level_2021} and NLI \citep{barker_ibm_2021} models for event coding. The prompting approach shares similarities with \citet{shin_constrained_2021}, who build a semantic parser to map natural text to canonical utterances. Their training set is constructed by prompting a language model in a human-in-the-loop fashion. \citet{DBLP:journals/corr/abs-2109-03659} uses NLI to extract relationship between two given entities based on a predefined hypothesis template.
\citet{schick_automatically_2020} present an approach to identify words that can serve as high-accuracy labels for text classification. However, they are not focusing on interpretability and a particular application domain such as political event coding. There also exist methods for automating prompt generation and selective incorporation of examples in the prompt \cite{shin_autoprompt_2020, gao_making_2021}. Existing work in prompt-based classification focuses on sentiment, topic or intent \cite{yin_benchmarking_2019, liu_pre-train_2021, schick_exploiting_2021}. 

Within the field of event coding, we distinguish work on event detection, event type ontologies, and automated event coding tools. Our work falls into the latter two. The World Event/Interaction Survey (WEIS) project \cite{mcclelland_world_1984} was pioneering in event data collection and event ontology design. The WEIS successor CAMEO \cite{schrodt_cameo_2012} is one of the most popular event ontologies until today and used by ICEWS \cite{boschee_icews_2015} and NAVCO \cite{lewis_nonviolent_2016} among others. VRA-Reader \cite{king_automated_2003} is among the first to automatize event coding based on matching string patterns. Its successors BBN ACCENT \cite{boschee_icews_2015}, Tabari and Petrarch2 \cite{norris_petrarch2_2017} rely on lambda calculus-based semantic parsing. Recent event coding systems rely on supervised machine learning \cite{hurriyetoglu_proceedings_2021, stoehr_classifying_2021, stoehr_ordinal_2022, stoehr_ordered_2023}, word embedding- \cite{kutuzov_tracing_2017, piskorski_tf-idf_2020} and transformer-based models \cite{olsson_text_2020, re_team_2021, hu_conflibert_2022, skorupa_parolin_multi-coped_2022}. 

\section{Conclusion}

We proposed a method to select answer candidates from prompts using textual entailment. This combined usage of state-of-the-art tools is motivated by a real-world use case that benefits humanitarian aid efforts with scarce resources. 

\vspace{1.75em}
\hspace{0em}\includegraphics[width=1.15em,height=1.15em]{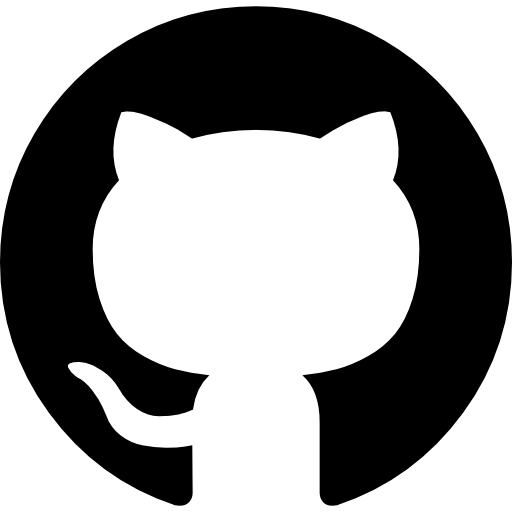}\hspace{.7em}\parbox{\dimexpr\linewidth-7\fboxsep-2\fboxrule}{\small \url{https://github.com/Clement-Lef/pr-ent}}

\section*{Acknowledgments}
This work was funded by the \href{https://eha.swiss/case-study/monitoring-patterns-of-violence/}{ETH4D Humanitarian Action Challenge} and grew out of a collaboration with Roberto Castello, Silvia Quarteroni, Daniel Gatica-Perez and Sandro Saitta. We would like to thank Fiona Terry, Francesca Grandi and Chiara Debenedetti from the International Committee of the Red Cross (ICRC) for feedback and discussions that motivated this research project. Niklas Stoehr is supported by a scholarship from the Swiss Data Science Center (SDSC).

\section*{Limitations}

We explore potential failure modes and the impact of bias in pre-trained (cloze) language models in \cref{sec:ablation}. Erroneous event coding can be further mitigated through incorporation of confidence score. In \cref{sec:limitations}, we discuss definitional caveats and model limitations. We make our code and interactive dashboard available for replication and scrutiny by the scientific community. We provide hyperparameter settings, training times and details on the computing infrastructure in the appendix (\cref{sec:reproducibility}). Since we are only considering off-the-shelf models, mostly without further fine-tuning, our experiments can be reproduced with limited computing resources. Our experiments are limited to English language, but can be extended by considering models pre-trained on other language data.

\section*{Impact Statement}

As explained in \cref{sec:intro}, our approach is aimed at helping low-resource organizations to analyze large amounts of text data efficiently. We do not foresee risk of misuse beyond the risks already introduced by conventional event coding pipelines. However, we would like to emphasize that the intended use of our approach is to gain additional, empirical insights for research and monitoring purposes, rather than for completely automatized decision-making. Application cases such as filtering event datasets are described in \cref{sec:limitations} and \cref{sec:actor_target_coding} .

\bibliography{references,more-refs}
\bibliographystyle{acl_natbib}

\newpage
\appendix

\section{Reproducibility Criteria}
\label{sec:reproducibility}

\subsection{Experimental Results}

\begin{enumerate}
    \item A clear description of the mathematical setting, algorithm, and/or model
    \begin{itemize}
        \item See Section \cref{sec:pipeline_description}
    \end{itemize}
    \item Submission of a zip file containing source code, with specification of all dependencies, including external libraries, or a link to such resources (while still anonymized) 
    \begin{itemize}
        \item Provided in the submission
    \end{itemize}
    \item Description of computing infrastructure used
    \begin{itemize}
        \item PR-ENT inference: Dell Latitude 7490 laptop - Intel(R) Core(TM) i7-8650U CPU @ 1.90GHz / 16 GB RAM
        \item DistilBERT finetuning: Macbook Pro M1 Max - M1 Max / 32 GB RAM
        \item Dashboard: 8 CPU Cores / 16 GB RAM
    \end{itemize}
    \item The average runtime for each model or algorithm (e.g., training, inference, etc.), or estimated energy cost
    \begin{itemize}
        \item Training:
        \begin{itemize}
            \item No training done for PR-ENT
            \item For comparison purposes, a DistilBERT model was fine-tuned on 3000 samples. It took several minutes on a laptop.
        \end{itemize}
        \item Inference:
        \begin{itemize}
            \item PR-ENT: 1-10secs per text depending on text length on a laptop
        \end{itemize}
    \end{itemize}
    \item Number of parameters in each model:
    \begin{itemize}
        \item DistilBERT-base-uncased (\href{https://huggingface.co/distilbert-base-uncased}{https://huggingface.co/distilbert-base-uncased}): 65M
        \item RoBERTa-large-mnli (\href{https://huggingface.co/roberta-large-mnli}{https://huggingface.co/roberta-large-mnli}): 125M
        \item RoBERTA-large-squad2 (\href{https://huggingface.co/deepset/roberta-large-squad2}{https://huggingface.co/deepset/roberta-large-squad2}): 125M
        \item PR-ENT: Top K, Entailment Threshold
    \end{itemize}
    \item Corresponding validation performance for each reported test result
    \begin{itemize}
        \item Not applicable
    \end{itemize}
    \item Explanation of evaluation metrics used, with links to code
    \begin{itemize}
        \item \href{https://scikit-learn.org/stable/modules/generated/sklearn.metrics.f1_score.html}{F1 Score, Scikit-learn}
        \item \href{https://scikit-learn.org/stable/modules/generated/sklearn.metrics.precision_score.html}{Precision, Scikit-learn}
        \item \href{https://scikit-learn.org/stable/modules/generated/sklearn.metrics.recall_score.html}{Recall, Scikit-learn}
        \item \href{https://scikit-learn.org/stable/modules/generated/sklearn.metrics.accuracy_score.html}{Accuracy, Scikit-learn}
        \item \href{https://docs.scipy.org/doc/scipy/reference/generated/scipy.spatial.distance.jensenshannon.html}{Jensen Shannon Distance, Scipy}
    \end{itemize}
\end{enumerate}

\subsection{Hyperparameter Search}

Not applicable

\subsection{Datasets}

\begin{enumerate}
    \item Relevant details such as languages, and number of examples and label distributions
    \begin{itemize}
        \item ACLED: See section \cref{sec:dataset_description}
        \item GTD: See section \cref{sec:dataset_description}
    \end{itemize}
    \item Details of train/validation/test splits
    \begin{itemize}
        \item ACLED: 3000 train sample / 1000 test sample
        \item GTD: 100 train sample / 1000 test sample
    \end{itemize}
    \item Explanation of any data that were excluded, and all pre-processing steps
    \begin{itemize}
        \item See section \cref{sec:dataset_description}
    \end{itemize}
    \item A zip file containing data or link to a downloadable version of the data
    \begin{itemize}
        \item ACLED: Data is not open source. We provide a json file containing the event ID used in train and test set.
        \item GTD: Data is available on \href{https://www.start.umd.edu/gtd/}{GTD Website} : We provide a json file containing the event ID used in train and test set
        \item Provided in the submission
    \end{itemize}
    \item For new data collected, a complete description of the data collection process, such as instructions to annotators and methods for quality control.
    \begin{itemize}
        \item Not applicable
    \end{itemize}

\end{enumerate}
\vspace{10cm}
\section{Additional Material}
\label{sec:appendix}

\begin{table*}[h]
\fontsize{10}{10}\selectfont
\centering
\renewcommand{\arraystretch}{1.5} 
\setlength{\tabcolsep}{0.5em} 
\begin{tabular}{ll}
\textbf{Event Description + Extracted Actor-Target}                                                                                                                                                                                                                                                                                                                                                                                                   & \textbf{Action}                                                        \\ \hline
\begin{tabular}[c]{@{}l@{}}Arrests: \textbf{{[}WHO (31\%): [LOC] police{]}} captured \textbf{{[}WHOM (90\%): [NAME]{]}}, \\ a senior [ORG] in [LOC]\end{tabular}                                                                                                                                                                                                                                                 & arrested
 \\ \cline{1-1}
\begin{tabular}[c]{@{}l@{}}On 3 January 2020, \textbf{{[}WHO (17\%): [LOC] Armed Forces{]}} regained [LOC], [LOC], \\ {[}LOC{]}, [LOC] and [LOC] from [ORG]. In the operations 6 [ORG] fighters were \\arrested and \textbf{{[}WHOM (67\%): 461 kidnapped civilians{]}} were rescued.\end{tabular}                                                                                                                      & rescued                                                   \\ \cline{1-1}
\begin{tabular}[c]{@{}l@{}}On 12 March 2020, \textbf{{[}WHO (40\%): police and military intelligence officers{]}} raided \\the home of retired \textbf{{[}WHOM (15\%, \underline{6\%): Lt. Gen [NAME]{]}}}. The candidate was arrested \\ and charged with treason in relation to remarks he made during a \underline{{[}WHO (29\%): TV{]}}\\ interview; his staff of 18, as well as the MP for [ORG] as well as his son have all been arrested.\end{tabular} & \begin{tabular}[c]{@{}l@{}}\textbf{arrested};\\ \underline{interviewed}\end{tabular}
\end{tabular}
\caption{Actor-target coding based on our pipeline augmented with an additional extractive question-answering (QA) model. The first example represents a clear ``who-did-what-to-whom'' pattern. In the second example, actor and target are separated into two sentences. Finally, the third example shows an event with two \emph{ARG0-V-ARG1} patterns (bolded and underlined). The confidence of the QA model is displayed for each answer.}
\label{tab:qa_prent_example}
\end{table*}

\subsection{Actor and Target Coding.}
\label{sec:actor_target_coding}

Until now, we studied how to code event types, which can be seen as actions or predicates of an event. We propose an extension to extract the actor and target of an event using question answering models similar to \citet{halterman_corpus-level_2021}. In \citet{he_question-answer_2015}, questions are constructed around a known action performed in an event. Given the example ``Military injured two civilians.'', \OurApproach yields ``injured'' as an action. Using this action, we can construct the questions ``Who was injured?'' and ``Who injured people?'' which are then fed to a QA model \href{https://huggingface.co/deepset/roberta-base-squad2}{RoBERTa-base-squad2} \cite{rajpurkar_squad_2016}. We present examples of extracted ``who-did-what-to-whom'' patterns in \cref{tab:qa_prent_example}. Actor-target coding is even harder to evaluate, as there can be multiple actions / targets / actors in an event description and the abstract mapping between manually annotated entity types (e.g. civilians) and verbatim mentions (e.g. demonstrators) is not known.

\begin{figure}[h]
     \centering
     \includegraphics[width=1\linewidth]{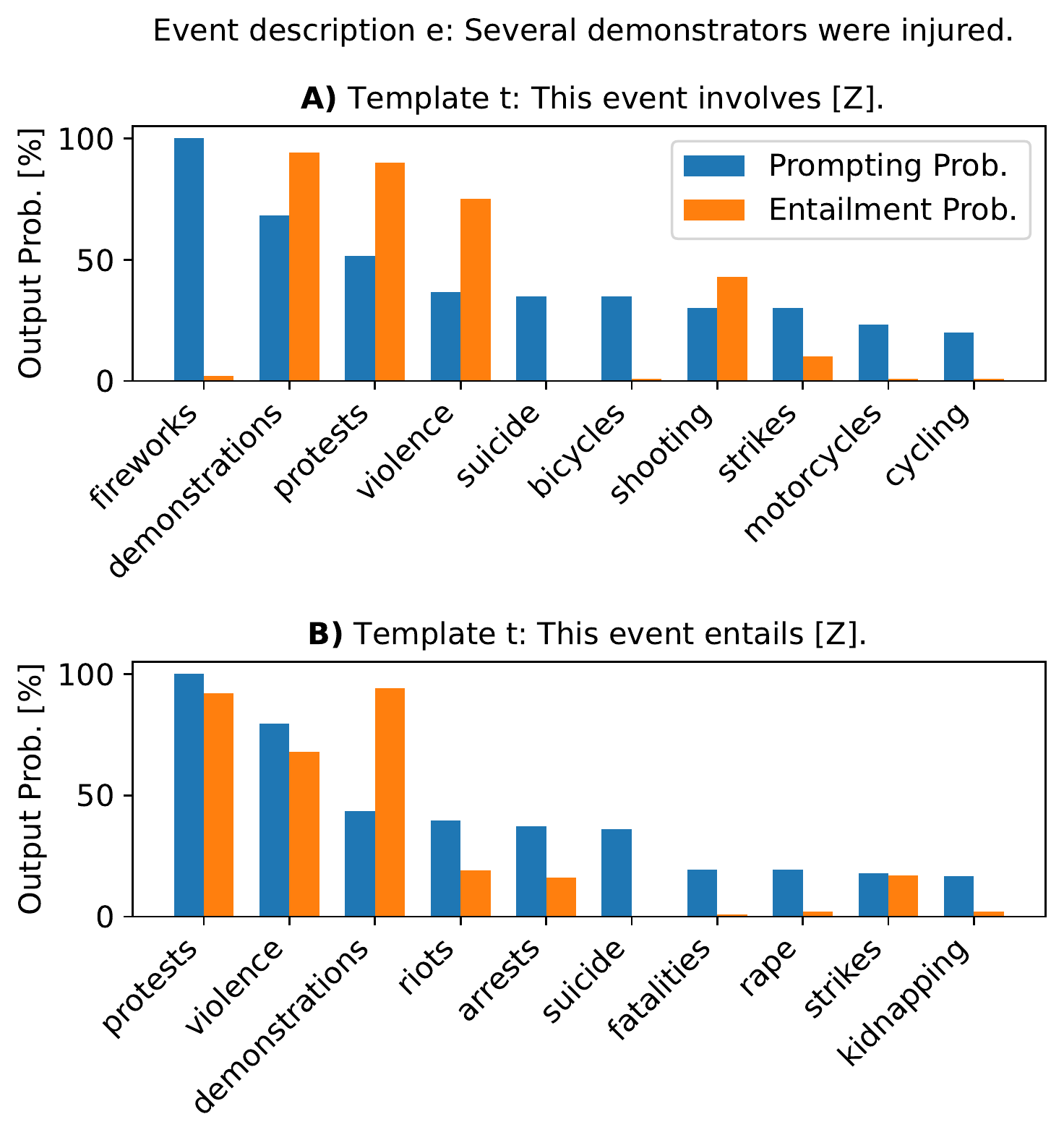} 
     \caption{Given the event description \emph{``Several demonstrators were injured.''}, and the two similar templates \textbf{(A)} and \textbf{(B)}, we get drastically different answer candidates as shown by the top 10 outputs of the prompt model (blue bar). However, in both cases we obtain the same \num{3} answer candidates if they are filtered through an additional entailment step (orange bar).
     }
     \label{fig:pnli_robustness}
 \end{figure}

\begin{table*}[h]
\fontsize{11}{11}\selectfont
\centering
\renewcommand{\arraystretch}{1.5} 
\setlength{\tabcolsep}{1em} 
\begin{tabular}{lll}
\textbf{Event Type}                  & \textbf{Template}                                                                           & \textbf{Entailed Answer Candidate}                                            \\ \hline
\multicolumn{1}{l|}{Arrest}          & People were {[}$\bZ${]}.                                                                        & arrested  \texttt{AND NOT} kidnapped                                                      \\ \hline
\multicolumn{1}{l|}{Killing}         & \begin{tabular}[c]{@{}l@{}}This event involves {[}$\bZ${]}.\\ People were {[}$\bZ${]}.\end{tabular} & \begin{tabular}[c]{@{}l@{}}killing\\ killed\end{tabular}                      \\ \hline
\multicolumn{1}{l|}{Looting}         & This event involves {[}$\bZ${]}.                                                               & looting \texttt{OR}
 theft \texttt{OR} robbery                                                   \\ \hline
\multicolumn{1}{l|}{Sexual Violence} & \begin{tabular}[c]{@{}l@{}}This event involves {[}$\bZ${]}.\\ People were {[}$\bZ${]}.\end{tabular} & \begin{tabular}[c]{@{}l@{}}rape\\ abused \texttt{OR} raped\end{tabular}                \\ \hline
\multicolumn{1}{l|}{Kidnapping}      & \begin{tabular}[c]{@{}l@{}}This event involves {[}$\bZ${]}.\\ People were {[}$\bZ${]}.\end{tabular} & \begin{tabular}[c]{@{}l@{}}kidnapping\\ kidnapped \texttt{OR} abducted\end{tabular}    \\ \hline
\multicolumn{1}{l|}{Protest}         & \begin{tabular}[c]{@{}l@{}}This event involves {[}$\bZ${]}.\\ People were {[}$\bZ${]}.\end{tabular} & \begin{tabular}[c]{@{}l@{}}protest \texttt{OR} demonstration\\ protesting\end{tabular}
\end{tabular}
\caption{Example of an event ontology designed by means of our approach of entailment-based prompt selection $\OurApproach$. The final ontology is defined in terms of templates and expected entailed answer candidates. We use the event type ``Killing'' versus all others to classify ``lethal'' versus ``non-lethal'' events in \cref{tab:lethal_event_results}. It's also used to compute results of \cref{fig:mali_kidnappings} and \cref{fig:ethiopia_protests}.}
\label{tab:ontology_from_prompt}
\end{table*}

\begin{figure*}[h]
 \centering
 \includegraphics[width=1\linewidth]{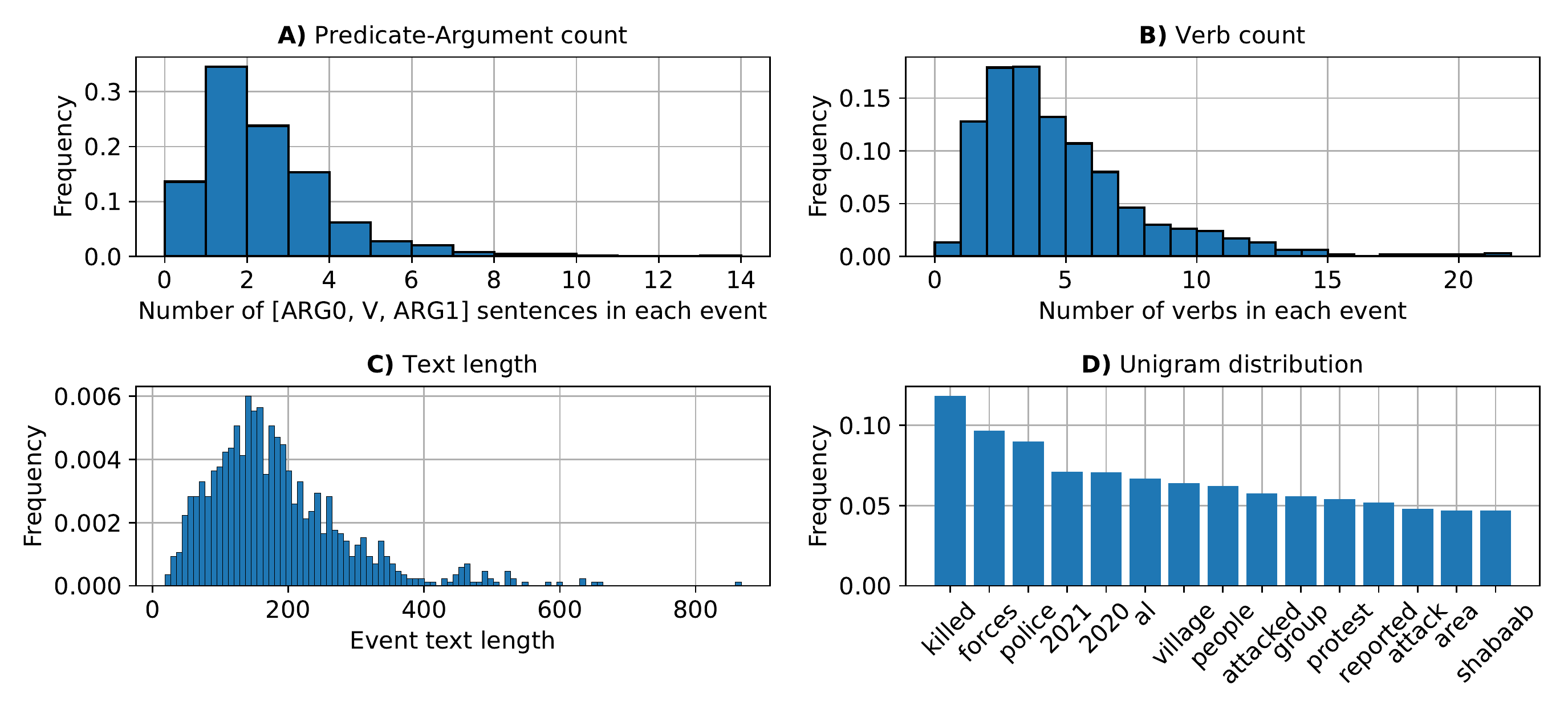} 
 \caption{Statistics over a sample of \num{1000} ACLED event descriptions; \textbf{(A)} encountering many predicate-argument structures per event description can be an indication of difficult event coding; \textbf{(B)} number of verbs (actions) per event description; \textbf{(C)} length distribution of event descriptions; \textbf{(D)} unigram distribution over dataset.}
 \label{fig:dataset_stats}
\end{figure*}

 \begin{figure*}[h]
     \centering
     \includegraphics[width=0.9\linewidth]{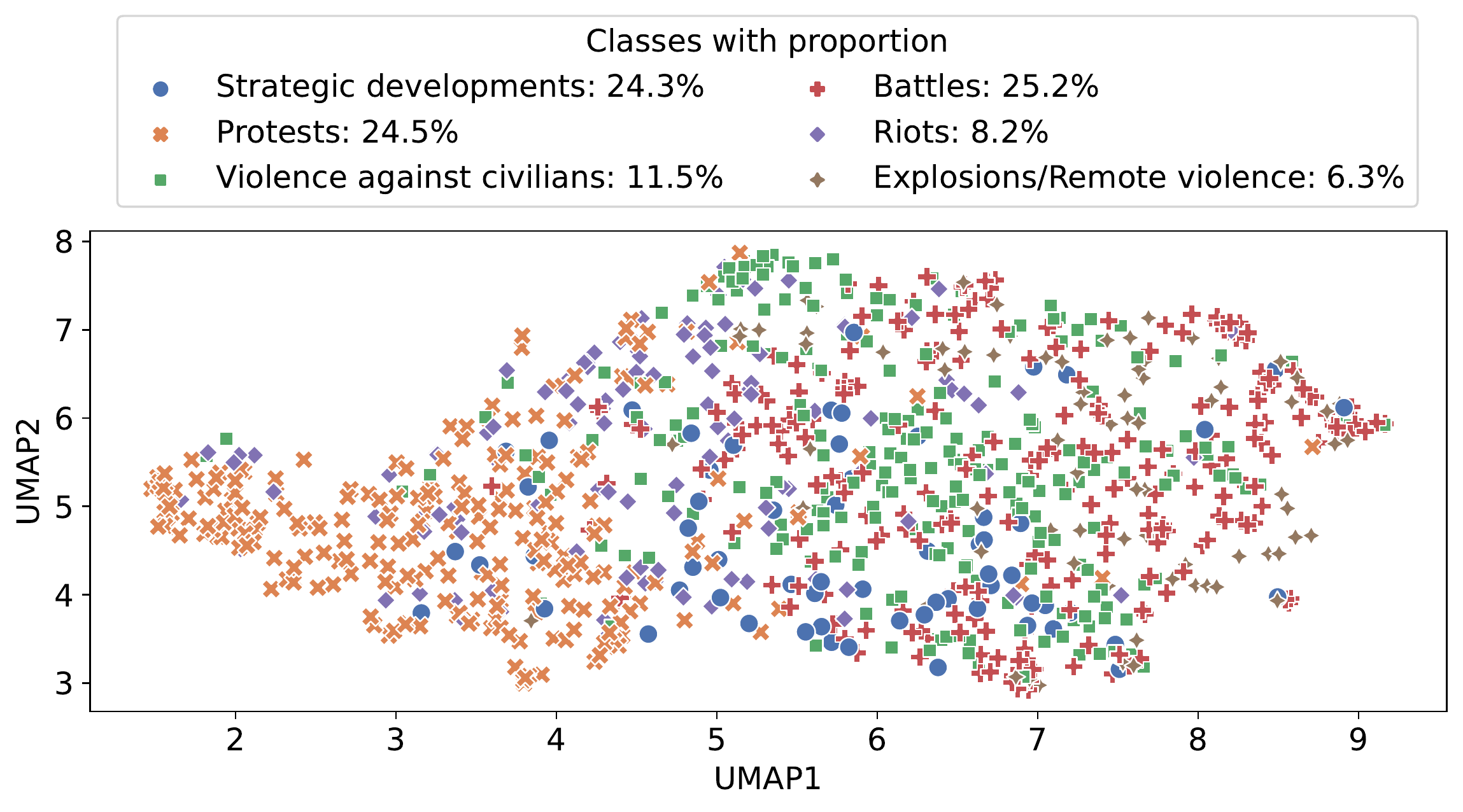} 
     \caption{Event type distribution as visualized using UMAP over GloVe embeddings of the event descriptions. While some event types are easily distinguishable from each other (e.g. \emph{Protests} and \emph{Battles}), others are harder to tell apart (e.g. \emph{Protests} and \emph{Riots}). We also show the proportion of each event type in the legend.
     }
     \label{fig:dataset_umap}
\end{figure*}

\begin{figure*}[h]
 \centering
 \includegraphics[width=1\linewidth]{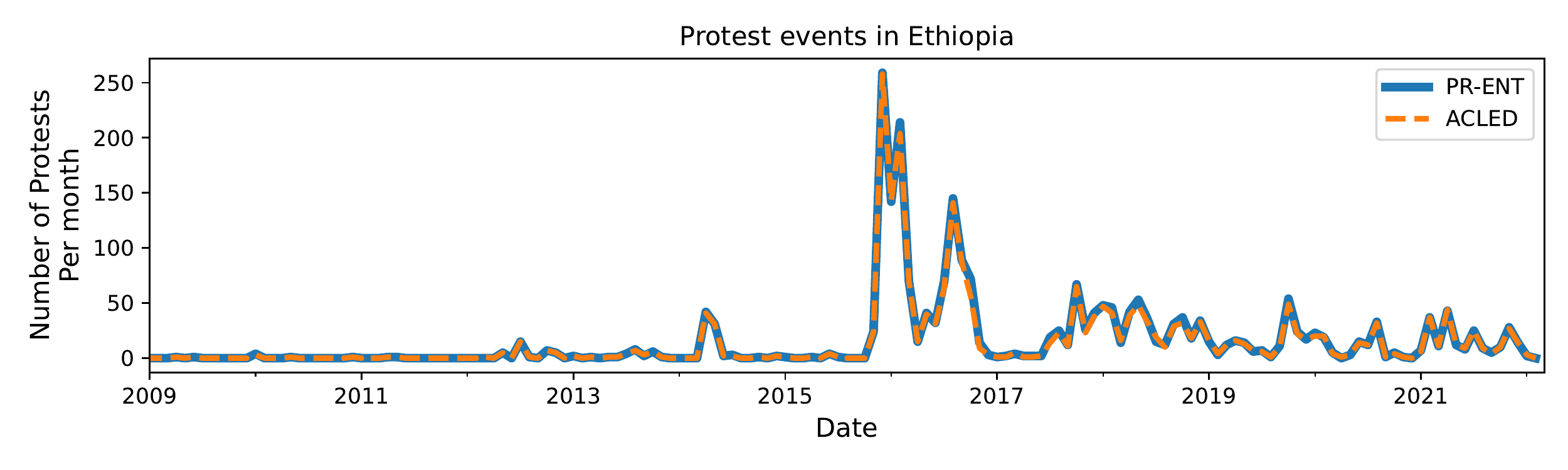} 
 \caption{Time series of the number of protest events per month in Ethiopia between 2009-2021. The dashed line corresponds to all protest events coded by ACLED annotators. The blue line corresponds to all protest events coded by \OurApproach. Despite \OurApproach codings being machine-automated, they are very similar to ACLED's codings. Both clearly detect the high intensity violence periods in 2016.
 }
 \label{fig:ethiopia_protests}
\end{figure*}

\begin{figure*}[h]
 \centering
 \includegraphics[width=1\linewidth]{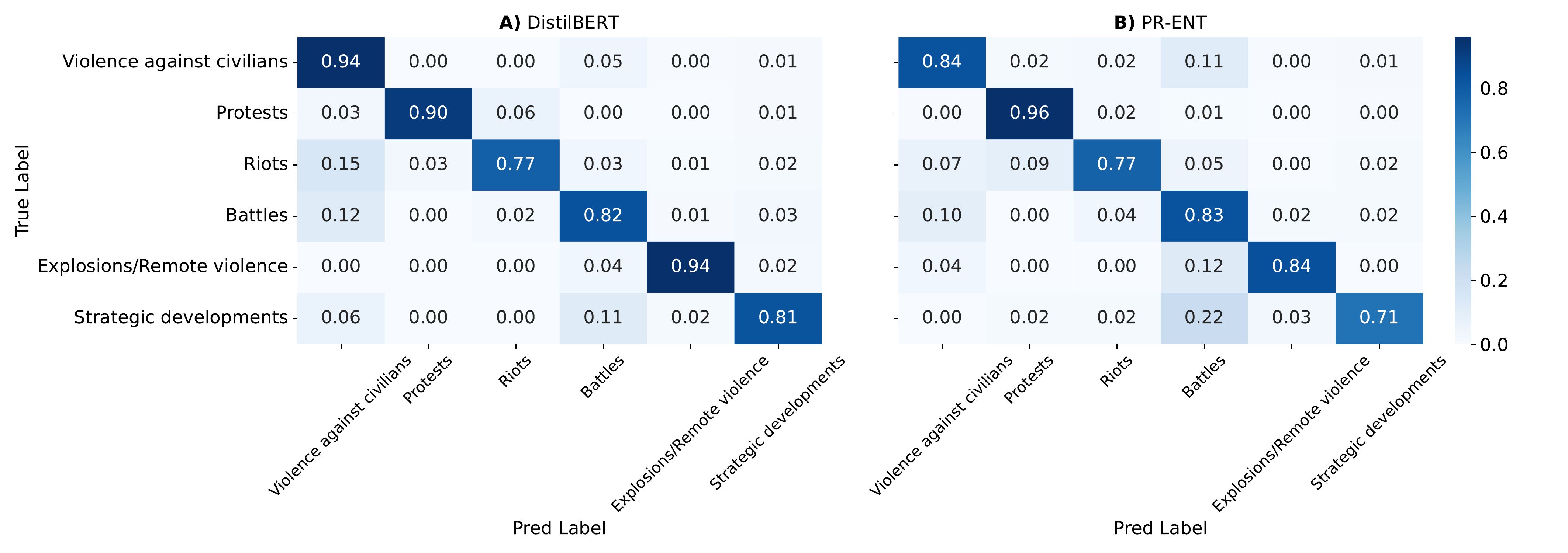} 
 \caption{Confusion matrices of DistilBERT and PR-ENT + LR on the test set.}
 \label{fig:cf_matrices}
\end{figure*}

\end{document}